# Analysis and Visualization of Linguistic Structures in Large Language Models: Neural Representations of Verb-Particle Constructions in BERT

Hassane Kissane[a], Achim Schilling[b, c], Patrick Krauss[b,c]

[a] Department of English and American Studies, University Erlangen-Nuremberg, Erlangen, Germany

[b] Neuroscience Lab, University Hospital Erlangen, Erlangen, Germany

[c] Pattern Recognition Lab, University Erlangen-Nuremberg, Erlangen, Germany

# Analysis and Visualization of Linguistic Structures in Large Language Models: Neural Representations of Verb-Particle Constructions in BERT

This study investigates the internal representations of verb-particle combinations within transformer-based large language models (LLMs), specifically examining how these models capture lexical and syntactic nuances at different neural network layers. Employing the BERT architecture, we analyse the representational efficacy of its layers for various verb-particle constructions such as "agree on", "come back", and "give up". Our methodology includes a detailed dataset preparation from the British National Corpus, followed by extensive model training and output analysis through techniques like multi-dimensional scaling (MDS) and generalized discrimination value (GDV) calculations. Results show that BERT's middle layers most effectively capture syntactic structures, with significant variability in representational accuracy across different verb categories. These findings challenge the conventional uniformity assumed in neural network processing of linguistic elements and suggest a complex interplay between network architecture and linguistic representation. Our research contributes to a better understanding of how deep learning models comprehend and process language, offering insights into the potential and limitations of current neural approaches to linguistic analysis. This study not only advances our knowledge in computational linguistics but also prompts further research into optimizing neural architectures for enhanced linguistic precision.

1. Introduction:

Recent neural network models have proven successful in most natural language processing tasks. This success is due to the large amounts of data on which neural network models are trained. These models have provided answers to several questions regarding language processing and have contributed to solving many issues, such as machine translation (MT) and automatic speech recognition (ASR). However, with the advent of neural network models, research questions have arisen regarding their abilities to deal with language. Questions such as: How much linguistic competence do neural network models have? What kind of linguistic knowledge is acquired by a neural network model? These questions aim to examine the linguistic abilities that neural networks possess, starting from phonological interactions, and syntactic tagging, to semantic and abstract representation.

A major area of interest has focused on the ability of deep neural networks to learn pre-defined linguistic concepts such as parts-of-speech tags and semantic tags (Dalvi et al., 2022). Some studies are attempting to explore in which part of the network specific linguistic knowledge is encoded. For example, Wilcox et al. (2018) reported that Recurrent Neural Networks (RNN) language models can learn about empty syntactic positions. Jawahar et al. (2019) investigated the linguistic structure learned by BERT: Pre-training of Deep Bidirectional Transformers (Devlin et al., 2019) and found that the language model can capture the rich hierarchy of linguistic information in the lower layers, including phrase-level information, syntactic features in the middle layers, and semantic features in the higher layers.

Several studies have been conducted to evaluate the capability of neural models to capture different types of semantic knowledge. Yet, after a short survey of the literature in the field, we found that not many studies interpret the state-of-the-art language models and explain their behaviour while processing specific and controversial linguistic assumptions, such as the nature of syntax and semantics of verb-particle combinations.

### *1.1. Linguistic Knowledge and the Distributional Models of Language:*

Linguistic knowledge or knowledge of language is a topic that has been approached mostly in early and late generative frameworks (Chomsky, 1968: p.ix; Chomsky, 1981) and, recently, in usage-based approaches such as construction grammar (Chaves, 2019; Leclercq, 2023). It is a term that refers to the ability of humans to "know a language," and this ability is different from the process of "having memorized a list of messages" (Rizzi, 2016). With this understanding of linguistic knowledge from the perspective of generative theory, this concept would refer to the grammar that includes the general principles and processes that allow the native speaker to construct and evaluate sentences in their language and decide on sentence acceptability. Therefore, some questions have been asked to answer the nature of linguistic knowledge: What is linguistic knowledge? How is it acquired? The second question focuses on how much of our language knowledge is decided by experience and how much is determined by a predefined mental process (Haegeman, 1994).

Linguistic knowledge is an umbrella term that includes components such as linguistic levels. When we talk about knowing a set of words of a certain language, that's "lexical knowledge". Knowing how those words are structured with each other and put together to form acceptable phrases or sentences is "syntactic knowledge". Knowing the meaning of words and their conceptual relations is "semantic knowledge." Mostly, communication in a foreign language is possible when knowing words and their meanings without knowing grammar, but not vice versa. Thus, it can be said that semantic knowledge precedes syntactic knowledge (Gärdenfors, 2017).

Semantic knowledge comprises the concepts, words, and categories stored in humans' semantic memory. It refers to the general knowledge of features that have been acquired and abstracted by humans from their experience. These concepts are comprised of semantic knowledge connected to words and stored in categories with other words that share the same semantic features (Kintz & Wright, 2016). It has been stated that categorization knowledge in humans has to be flexible. For instance, when a child learns a word, the other words that belong to the same semantic domain will be learned much faster (Gärdenfors, 2017).

Building on the concept of the distributional hypothesis (Harris, 1954), where words derive meaning from their context of use and contrasting relationships with other words, distributional models emerge as computational techniques inspired by these principles. These models analyse the statistical distribution of words in large amounts of text data, focusing on how frequently words co-occur with others. This co-occurrence analysis aims to capture the semantic meaning of words based on their surrounding context, similar to how distributional semantics emphasises the role of contrasts. Thus, distributional models act as computational tools for representing and analysing linguistic knowledge.

Distributional semantics is primarily derived from structuralist traditions (Sahlgren, 2008). As in structuralism, words are defined by their location in a system, the lexicon, based on a set of features; their values are established by contrasts in the contexts of the usage of words (Boleda, 2020). Since the emergence of distributional models of language, there has been a growing trend in the cognitive sciences to support the connection between embodied cognition and distributional approaches (Barsalou et al., 2008; Lenci, 2008).

This technique produces semantic representations as a by-product of language prediction tasks. For example, Mikolov, Yih, and Zweig (2013) employed language modelling, which is the function of predicting words in sentences. In a predictive setup, word vectors (also known as embeddings) are generally initialised at random and repeatedly adjusted as the model traverses the data and improves its predictions. Because comparable words exist in similar situations, they have similar embeddings, which are part of the model's internal representation for predicting a certain phrase (Boleda & Herbelot, 2016). Based on the same principle of training on certain tasks, but with contextual embedding, Devlin et al. (2019) produced the pre-trained BERT model on a large amount of data with two false tasks: masked language modeling and next sentence prediction. Therefore, the pre-trained model can provide embedding of the newly provided input by only passing the text through it.

The early models of distributional semantics are significant components in understanding semantic knowledge and how meaning is derived from context, which means they do not directly address other aspects of linguistic knowledge (or they are not designed to) such as syntax or morphology. However, recent neural language models with their advanced contextualized embedding techniques like BERT, which are inspired by the distributional hypothesis, can be used in computational linguistics to analyze various aspects of language beyond just semantics. For instance, the main goal of distributional models is to define the similarity/dissimilarity between the co-occurrences of the word "give" in the sentences *She gave up the job* and *She gave her van a scratch*. Moreover, further analysis of the differences between the representation of these different contexts in the internal activation of the models allows investigation of the different linguistic nature (syntax) of the word "give" in the two contexts, like the transitivity of the verb in the first given context and the ditransitivity in the second context. Accordingly, it was proposed by Wingfield et al. (2022) that linguistic distributional models represent a cognitively plausible approach to understanding linguistic distributional knowledge when they are assumed to represent an essential component of semantics.

### 1.2. Linguistic Levels and Corresponding BERT layers:

The model's **Initial Layers** usually process the sequence form. Rogers et al. (2020) state that the first BERT layer processes the combination of token, segment, and positional embeddings. Early layers to the fourth layer in the BERT-base, contain more information about linear word order. However, as the processing moves up the layers, an emergent understanding of more hierarchical sentence structure appears.

**Middle Layers** are specified in syntactic information processing. Studies by Hewitt and Manning (2019), Goldberg (2019), and others have found these layers (e.g., layers 6-9 in BERT-base and 14-19 in BERT-large) most effective in capturing syntactic structures, like tree depth and subject-verb agreement. Jawahar et al. (2019) have approved the findings in these studies, which have shown that phrasal representations learned by BERT in the middle encode a rich hierarchy of linguistic information.

**Late Layers:** During fine-tuning for specific tasks, these layers undergo the most change. However, restoring lower layers to their pre-trained states does not significantly affect model performance, indicating a certain robustness.

### 1.3. Internal activations of the model:

Within natural language processing and neural models' interpretation, BERT represents one of the state-of-the-art transformer-based language models. The success of this model in numerous NLP tasks has prompted ongoing research into understanding its predictions and examining the knowledge it embodies (Manning et al., 2020; Rogers et al., 2020; Belinkov, 2022; Oh & Schuler, 2023). Unlike the traditional embedding systems (Mikolov et al., 2013; Pennington et al., 2014), the output of a single layer in BERT, which is named the "hidden states" is contextualised, aligning with the principle of mutual information maximization. This context-based representation implies that every token within an input sentence to the model is represented by a unique vector for the token and its occurrence in the specific context (Miaschi and Dell'Orletta 2020). Therefore, it is reported that "distilled contextualised embeddings better encode lexical semantic information" (Rogers et al., 2020).

The contextual embeddings provided by the BERT model could be used to answer questions about the model's abilities to represent complex constructions like those tokens that are involved in multi-word constructions. For instance, the token "agree" in the prepositional constructions "agree on, agree with" and the token "give" in the phrasal verbs give up, give out, give in". Thus, a further way to learn about a language model's internal activations is to extract the hidden states that each Transformer block produces, representing these context-sensitive constructions to explain the nature of linguistic unit representations in a neural language model.

Supporting these questions, Wiedemann et al. (2019), Schmidt and Hofmann (2020) have found evidence of distinct clusters in BERT embeddings that correspond to different word senses. However, Mickus et al. (2019) point out that the representations of the same word can vary depending on the word's position within a sentence, highlighting a dynamic aspect of BERT's model in capturing linguistic differences. This evolving understanding of BERT embeddings provides a background for further exploration, particularly in interpreting neural language models.

## 1.4. Construction Grammar as a Theoretical Framework for Neural Language Models Interpretability:

Recent developments in cognitive-functional linguistics and usage-based linguistics suggest that language structure emerges from language use, with the essence of language being its symbolic dimension, with grammar being derivative (Tomasello 2003: 5). Construction grammar is one of the usage-based language acquisition approaches (Perek, 2023). From this point, it provides a corpus-based analysis of the process of language acquisition; which means there is no assumption that the children have an innate language faculty to process and acquire language. instead, they acquire language because they are living in a linguistic community. From this principle, grammar would be considered a dynamic system of form-meaning pairings, that is based on the domain-general acquisition by children, in which the linguistic categories emerge by processing large amounts of linguistic data (Diessel, 2013).

Constructional approaches to language description have their roots in Fillmore's early work from 1968. From there, a variety of approaches emerged with the shared proposals that (a) grammar is directly related to function (grammar has meaning) and does not depend on transformations and derivation (structure), and (b) constructions are learned pairings of form and meaning related to one another in a constructional network, in which the relationships between and among constructions are captured via a "default inheritance network" (Goldberg, 2013).

From this principle, construction grammar shares the same principles with embedding-based models that represent linguistic knowledge as vectors in high dimensional space which they are trained on large corpora to handle linguistic tasks. Therefore, the understanding of linguistic structures in those models emerges from language use, in which the representation of constructions are statistical distributions with quantitative information derived from corpus data (Rambelli et al, 2019). As experience shapes linguistic knowledge, distributional characteristics become an important factor in determining the content of linguistic representations. In this perspective, language is not regarded as an autonomous cognitive system. Rather, grammar acquisition is treated as any other conceptualization process, and language knowledge evolves through usage. (Pannitto & Herbelot, 2023: 28).

Although, Lexical units may not capture all information contained in larger linguistic units (sentences) as stated in Goldberg, (1995), which aligned with the constructional aspect of the integration of syntax with semantics in terms of that syntax has meaning. With distributional models of language, it is supposed that the vector space of the lexical unit has much information about the structure of the sentence it appears in.

In commitment to the constructionist approach while attempting to make the fewest feasible assumptions on the actual content of linguistic knowledge, it is speculated that language is a network of structures that are meant to approximate construction. Because constructions are form-meaning pairs, the concept of grammar encompasses a meaning space that extends beyond the lexical level. This is readily done by expanding the concept of vector space models, which has been extensively researched and applied in distributional semantics (Pannitto and Herbelot, 2023; 25) where the linguistic elements (mainly lexical entries) are represented in high dimensional space and organized in groups according to sets of relations that are derived from the linguistic context they appear in.

      Several studies have been done on the plausibility of constructional approach to language, some of them are neurolinguistic (Arbib & Lee, 2008; Thompson et al, 2007) and some others are computational models that used neural language models. Earlier, a study has done on computational modeling of the constructivist approach to language

development, introducing a construction grammar formalism for emergent grammars. It proposes invention, abduction, and induction as necessary for language learning. Experimental results have been obtained with agents playing situated language games, but more technical work is needed. This Early approach is called Fluid Construction Grammar (steels, 2004; Steels and De Beule, 2006; Steels, 2011, 2017; van Trijp et al., 2022).

One other aspect of the alignment between construction grammar and neural language models is that syntax is integrated with semantics. In contrast to formal approaches such as generative grammar framework which has long defended the idea that the study of syntax is a separate entity from other aspects of language (see, e.g., Chomsky 1957, piantadosi 2023). Neural language models encode syntax with semantics in their output representations. This behaviour of neural language models is entirely suitable with the core idea of construction grammar that sees constructions as form-meaning pairings in which the main linguistic unit (the construction) in the constructionist approach could not be analysed in a separation between the form and the meaning. Models encode words as vectors in a high-dimensional space and can predict syntactic properties such as part of speech tags (PoS) without separating them from semantic representations or analysing other levels. This is achieved by encoding semantic properties into vectors and initializing word vectors via distribution semantics (Piantadosi, 2023). Madabushi et al, (2020) have approached the neural model (BERT) asking questions regarding the representation of constructional information within the neural network: (a) How does the addition of constructional information affect BERT? (b) How effective is BERT in identifying constructions? Their approach found that the neural network has access to significant constructional information, which is not explicitly available in the output layer but can be accessed within its internal activations. This observation demonstrates deep learning methods' capabilities and lexico-semantic information's ability to learn constructional information due to the redundancy inherent in language. A recent study by Weissweiler et al., (2023) found out that Construction Grammar Provides Unique Insight into Neural Language Models (LMs). They investigated the abilities of LMs to represent constructions and constructional information. Therefore, they argued that understanding constructions is essential for effective LMs.

### *1.5. Linguistic observations of the defined constructions:*

How many lexical units are considered while dealing with verb-particle combinations? The answer to this question is unclear. We might declare that any attested combination of traits represents a new sense, or we could pick a few features and say that specified combinations belong to unique senses. Given the feature space, both selections are somewhat arbitrarily made, and this suggests that the theory does not regard unique senses or lexical items as first-class language structures. This might be a result of the contextual modulation principle.

#### 1.5.1. *Agree on, Agree with, Agree to,* and *Agree that*:

The verb "agree" serves as the main lexical unit, its meaning is emerged from its form, not modified by the following prepositions or conjunctions. Thus, it has a one lexeme that it belongs to when it is sorted in an English dictionary with only one meaning, and the following prepositions or conjunctions are related with the verb in a syntactic relation depending on the phrase structure requirements. Since the verb "agree" is a single lexical unit, its meaning is not influenced by the preposition or conjunction that follows it. Moreover, it might have a single representation in NLP systems when it is investigated in as a specific token in an NLP model.

#### 1.5.2. *Come in, Come out* and *Come back*:

The verb "come" is combined with different adverbial particles (*in*, *out*, *back*) to form phrasal verbs with distinct meanings. Thus, the full combination is considered a dependent form-meaning construction, which should be sorted as a single lexeme in the English dictionary based on its meaning. Each combination of "come" with a different adverbial particle forms a distinct phrasal verb with its meaning. Therefore, since the NLP models are context-sensitive, the verb "come" may have multiple representations in NLP systems, each indicating the specific adverb that follows it. For example, *come_in*, *come_out*, *come_back*. However, the representations may appear similar in some ways, regarding the degree of semantic transparency of the phrasal verbs in this category; where the combination might correspond to the composition of the verb and the following particle which is named literal phrasal verbs.

**1.5.3.** *Give in, Give out, Give up*:

The verb "give" forms phrasal verbs with different particles (*in, out, up*), each standing for a unique meaning. Similar to "come", each combination of "give" with a different particle forms a distinct phrasal verb with its meaning. Thus, the different forms of the verb "give" may have more distinct representations in NLP systems because of its idiomatic nature, each indicating the specific particle that follows it. For instance, the following verbs *give_in*, *give_out*, and *give_up* have different meanings and have no participation of the verb or the particle in the meaning of the full combination.

In summary, while "agree" might have a single representation in NLP systems with features indicating the accompanying preposition or conjunction, "come" and "give" may have multiple representations, each reflecting the specific adverbial particle that modifies their meaning. This distinction arises from the different nature of these constructions: "agree" operates more as a unitary verb with different argument structures, while "come" and "give" form phrasal verbs where the combination of verb and particle determine single units of meaning.

### *1.6. Constructional Approach to verb-particle combinations:*

Lipka (1972) defined the category of phrasal verbs, which he calls "verb-particle construction" as "in English (and also in German) can be regarded as a particular surface structure shared by a large number of lexical items with various word-formative and semantic structures." In simpler terms, this definition says that phrasal verbs in English (and German) follow a recognizable pattern in their construction (verb + particle). Despite having a wide range of combinations and meanings, they share a common format that makes them identifiable as phrasal verbs. This pattern includes the way they are formed (verb combined with a particle) and their often-idiomatic meanings.

Within the framework of Lexical-Functional Grammar (LFG), a lexicalist, constraint-based grammatical approach that shares a lot of fundamental tenets with Construction Grammar. (CxG) (Findlay, 2023), and Booij (2001) argue against the morphological interpretation of phrasal verbs (called them Separable Complex Verbs (SCVs)), instead positioning them as syntactic constructs. He proposed that phrasal verbs are originated from resultative small clauses, but undergo grammaticalization. This process transformed the particles from clause elements into productive aspectual markers. The development of phrasal verbs exemplifies how syntactic surface structures can transcend their role as mere outputs of syntactic rules. They can evolve into their entities, ultimately resembling lexical idioms with open slots.

Boas (2003: 32, 236, 280), in his work on resultatives, focuses on the resultative aspect of the phrasal verbs, without taking into consideration the word order alternation. He provides the idea that the particle acts as the modifier of the preceding verb and is part of its semantic frame, but it is not part of the verb's syntactic frame. In his approach, the particle and verb are considered separated and co-occurred constructions, rather than as a syntactic unit.

Another observation of the verb-particle combinations is their idiomatic status, which means that the full combination meaning is not predicted from its parts. Thus, the verbs that act as predicates exhibit specificity in their usage in which the observation that the meanings resulting from their combinations are often not predictable from the individual components used in different contexts. For instance, the expression "take it off" can signify undressing, whereas "take it on" does not mean the opposite meaning (in contrast to "put it on"). When used intransitively, "take up with someone" typically denotes the commencement of a romantic relationship, while its transitive form, "take something up with someone", usually signifies the initiation of a conflict. "Take something over to" generally means to transport something, while "take someone down" conveys the notion of overcoming, and "take someone out" can suggest either a social outing or a violent act, such as murder (e.g., Goldberg, 2016; 2019: 57-58).

Within different approaches to the lexical and syntactic approaches to phrasal verbs, lexical methodologies align well with the accumulated neurobiological findings. Regarding phrasal verbs, they appear to be more suitable than the phrasal proposals posited by Booij (2001, 2002). The neurobiological plausibility for the lexical approaches to the phrasal verbs is presented by Cappelle et al. (2010: 197) The results based on the multi-feature Mismatch Negativity (MMN) design, indicated that phrasal verbs are considered as lexical units. It is evident that the increased level of activation observed for real phrasal verbs, in contrast to pseudo-verb-particle combinations, implies that a verb and its particle created a unified lexical representation, known as a single lexeme. Nevertheless, the inquiry remains unresolved on the essence of a discontinuous lexical unit. The concept of discontinuous words has a long history (Wells 1947), yet formal explications of this concept are not provided yet. (Müller, 2015: 658).

One of the cases that should receive attention in this context is the case of verb-preposition combination so-called by Quirk, Greenbaum, Leech, and Svartvik (1985) "Prepositional verb", in which the full combination such as (agree on, agree with) is considered a single lexical unit in Cruse's[1] definition. However, the category verb-preposition combination has received an alternative analysis within valency theory, which treats it differentially from phrasal verbs. The proposed analysis by Herbst and Schüller (2008: 120) is as follows:

| Model of Analysis | | | |
|---|---|---|---|
| Quirk et al. 1985 | | Herbst and Schüller (2008) | |
| verb | complement | verb | complement |
| *agree* | *to come early* | *agree* | *to come early* |
| *agree on* | *a common approach* | | *on a common approach* |
| *agree with* | *the last bit* | | *with the last bit* |

**Table 2:** Descriptive grammar versus valency approach to the analysis of verb-particle combinations

Therefore, this theoretical conflict needs an experimental design to explore the lexical and syntactic status of both phrasal verbs and verb-preposition combinations on whether they are both processed as unified lexical units which is the case of phrasal verbs as approved by neurolinguistic studies (Cappelle et al. (2010) or the verb-preposition

---

[1] Cruse writes on the definition of lexical unit: "A lexical unit is then the union of a lexical form and a single sense" (Cruse, 1986).

combinations are considered a different type of composed verbs and particles. Therefore, an analysis of the samples is provided following the constructional framework provided by Herbst & Hoffman (2024):

| DITRANSITIVE CONSTRUCTION (OBJ: TO-INF) | | |
|---|---|---|
| ÆFFECTOR shows their willingness to carry out a particular action [ÆFFECTED-ACTION]. | | |
| ÆFFECTOR | Action | ÆFFECTED-ACTION |
| SUBJ | V | Obj |
| They agreed to meet | | |

**Table 3**: The English DITRANSITIVE Construction.

| ON-SPECIFIC-ISSUE CONSTRUCTION | | |
|---|---|---|
| An AGENT performs an action with respect to a SPECIFIC-ISSUE. | | |
| ÆFFECTOR | | SPECIFIC ISSUE |
| SUBJ | V | OBJ: $PP_{on}$ |
| Once you agree on the level of quality, the price almost … | | |

**Table 4**: The English ON-SPECIFIC-ISSUE Construction.

| WITH-PARTNER CONSTRUCTION | | |
|---|---|---|
| An ÆFFECTOR performs an action together with a PARTNER. | | |
| ÆFFECTOR | | PARTNER |
| SUBJ | V | OBJ: $PP_{with}$ |
| I don't agree with the theory that my becoming emotionally charged up… | | |

**Table 5**: The English WITH-PARTNER Construction.

| MONOTRANSITIVE CONSTRUCTION (THAT-CLAUSE OBJ) | | |
|---|---|---|
| An ÆFFECTOR does something to an ÆFFECTED. | | |
| ÆFFECTOR | | ÆFFECTED |
| SUBJ | V | Obj: that-CL |
| The Minister agree that the loans scheme has turned out to be an administrative nightmare. | | |

Table 6: The English THAT-CLAUSE Construction.

| COME-IN/OUT/BACK-INTRANSITIVE-VERB-PARTICLE-CONSTRUCTION | | |
|---|---|---|
| An ÆFFECTOR carries out an action. | | |
| ÆFFECTOR | Action | |
| SUBJ | V | part |
| NP | COME | IN/OUT/BACK |
| Can I come in now? She felt that her teeth would come out altogether if she tried to bite Fleury's cake. Are you going to come back tomorrow? | | |

Table 7: The English COME-IN/OUT/BACK Verb-Particle Constructions.

| GIVE-AWAY-VERB-PARTICLE-CONSTRUCTION | | | |
|---|---|---|---|
| An ÆFFECTOR voluntarily transfers possession or control of something to another entity, often without expecting compensation. | | | |
| ÆFFECTOR | | | ÆFFECTED-ACTION |
| SUBJ | V | part | OBJ |
| or you could give away a free kick | | | |

Table 8: The English GIVE-AWAY Verb Particle Construction.

| GIVE-UP-VERB-PARTICLE-CONSTRUCTION | | | |
|---|---|---|---|
| An ÆFFECTOR decides not to continue something they have been doing for quite some time or not to pursue an action they had planned any longer. | | | |
| ÆFFECTOR | | | ÆFFECTED-ACTION |
| SUBJ | V | part | obj |
| NP | GIVE | UP | NP (≠ pron) V-ing-cl |

Table 9: The English GIVE-UP Verb Particle Construction.

| GIVE-OUT-VERB-PARTICLE-CONSTRUCTION | | | |
|---|---|---|---|
| An ÆFFECTOR distributes items to multiple recipients or reaches a point of depletion or cessation of function. | | | |
| ÆFFECTOR | | | ÆFFECTED-ACTION |
| SUBJ | V | part | obj |
| NP | GIVE | OUT | NP (≠ pron) V-ing-cl |

Table 10: The English GIVE-OUT Verb Particle Construction.

| GIVE-IN-INTRANSITIVE-VERB-PARTICLE-CONSTRUCTION | | |
|---|---|---|
| An ÆFFECTOR carries out an action. | | |
| ÆFFECTOR | Action | |
| SUBJ | V | part |
| NP | GIVE | IN |

Table 11: The English GIVE-IN Verb Particle Construction.

*1.7. **This Study:***

Despite their successes in natural language processing and other areas of artificial intelligence, recent deep neural networks remain difficult to explain and are still considered to be black-box models (El Zini & Awad, 2022). Here, we examine the internal representations created by transformer-based models at different levels of the network and assess their language expertise using appropriate extrinsic linguistic constructs. Following Belinkov et al. (2020), regarding the interpretation of neural models and linguistic-level representations within their internal activations and in connection with the aim of our study, which is the examination of the representation of verb-particle combination, we seek answers to the following questions: (i) Do the internal representations capture lexical semantics? (ii) How accurately is the phrase-level structure captured within the internal representations of individual words? Which layers in the architecture capture each of these linguistic phenomena? (iii) Can the internal activations be explained and meet the usage-based theories about verb-particle combinations like the unity of phrasal verbs and the syntactic compositionality of prepositional verbs?

## 2. Methods:

### 2.1. Dataset creation and pre-processing:

We collected natural language text data for training our model from the British national corpus, with queries to search the target construction (target verb + target particle) preceded by 10 tokens and followed by the other 10 tokens. The complete text data consists of a total of 995 samples. The number of samples representing each specific verb construction is (*agree on: 100*; *agree to*: 100; *agree that: 100*; *agree with*: 100; *come back*: 99; *come in*: 99; *come out*: 99; *give in*: 99; *give out*: 93; *give up*: 100, give away: 100).

The data pre-processing stage is critical for every successful model. It seeks to prepare the data for subsequent use. After selecting the principal dataset, we began the purification process on the selected data. We first imported the 're' library for these operations, and for each functionality, we defined a method to solve it. We have four functions: Removes punctuation from the sentence using the translation table created earlier (table). Remove leading and trailing whitespaces (e.g., spaces, tabs, and newlines). Replace multiple consecutive spaces with a single space. Converts the sentence to lowercase, making all characters in the sentence lowercase.

| Character | Pre-processing step |
| --- | --- |
| Punctuation (!"#$%&'()*+,-./:;<=>?@[\]^_`{|}~) | Removed |
| Leading and trailing whitespaces | Removed |
| Extra whitespaces | Replaced with single space |
| Uppercased characters | Lowercased |

**Table12:** Data cleaning. Words, characters and their replacements during data cleaning.

## 2.2. BERT architecture:

This section introduces (BERT) Bidirectional Encoder Representations from Transformers (Devlin et al., 2019) and its implementation for the study. BERT is a transformer-based model, following the architecture proposed by Vaswani et al. (2017). The BERT model employed a transformer design with 12 layers for the base model and 24 for the large model. Each block in the base model has 12 attention heads, while the large model has 16 attention heads and a hidden dimension of 768 in the base and 1024 in the large model.

BERT was pre-trained on the BooksCorpus (800 million words) (Zhu et al., 2015) and English Wikipedia (2,500 million words)[2]. Unlike the conventional left-to-right or right-to-left language models used by Peters et al. (2018a) and Radford et al. (2018), BERT was pre-trained using two unsupervised tasks. The first is the "masked LM," in which 15% of all WordPiece tokens in each sequence are randomly masked and replaced with the [MASK] token. The model then predicts the masked tokens, training a deep bidirectional representation. This process is often referred to as a Cloze task in the literature (Taylor, 1953). The second pre-training task is the next sentence prediction. To train a model that understands the relationships between two sentences, each pretraining example selects sentences A and B from the corpus. In 50% of cases, sentence B is the actual next sentence following A (labelled as IsNext), and in the other 50%, it represents a random sentence from the corpus (labelled as NotNext).

In this study, we use the original BERT[3], mentioned above. In addition to the constructional BERT (CxG-BERT) provided by Madabushi et al (2020). While we use the original pre-trained model without any further training or fine-tuning, the CxG-BERT is following the same structure of the original model, but with extra-training using sentences instantiating constructions that have a frequency from 2 to 10,000 instances.

### 2.3. Multi-dimensional scaling:

---

[2] Both the original BERT and CxG-BERT include large amount of text data from Wikipedia, which might be not standard from a pure linguist view. However, these models stand as main resources and are considered as the only available models for this kind of experimental design.

The neural model output for each target token is a vector representation of 768 dimensions, which is not visualizable. Therefore, we used a frequently used statistical technique for generating low-dimensional data from the high-dimensional embeddings. This method is multi-dimensional scaling (MDS) (Torgerson, 1952; Cox & Cox, 2008). In the context of our study, we use it to represent the selected constructions in a dataset as points in a multidimensional space so that close similarity between constructions in the dataset corresponds to close distances between the corresponding points in the representation. MDS is a productive implanting strategy to visualise high-dimensional point clouds by anticipating them onto a 2-dimensional plane. Moreover, MDS has the definitive advantage that it is parameter-free and all shared separations of the focuses are protected, in this manner preserving both the global and local structure of the primary data (Surendra et al., 2023). MDS is an effective approach for visualising high-dimensional data by representing patterns as points in space and dissimilarities as distances between points.

The data representation may be visualised as a series of point clusters by colour-coding each projected data point in a data set based on its label. For example, MDS has already been used to visualize word class distributions in various linguistic corpora (Schilling et al., 2021a), hidden layer representations (embeddings) of artificial neural networks (Schilling et al., 2021b; Krauss et al., 2021), the structure and dynamics of recurrent neural networks (Krauss et al., 2021), brain activity patterns during pure tone or speech perception (Schilling et al., 2021; Krauss et al., 2018), or even during sleep (Krauss et al., 2018).

### 2.4. Generalized discrimination value:

We used the generalized discrimination value (GDV) to calculate cluster separability as published and explained in detail in Schilling et al, (2021). Briefly, we consider $N$ points $\mathbf{x_{n=1..N}} = (x_{n,1}, \cdots, x_{n,D})$, distributed within $D$-dimensional space. A label $l_n$ assigns each point to one of $L$ distinct classes $C_{l=1..L}$. In order to become invariant against scaling and translation, each dimension is separately z-scored and, for later convenience, multiplied with $\frac{1}{2}$:

$$s_{n,d} = \frac{1}{2} \cdot \frac{x_{n,d} - \mu_d}{\sigma_d}.$$

Here, $\mu_d = \frac{1}{N}\sum_{n=1}^{N} x_{n,d}$ denotes the mean, and $\sigma_d = \sqrt{\frac{1}{N}\sum_{n=1}^{N}(x_{n,d} - \mu_d)^2}$ the standard deviation of dimension d. Based on the re-scaled data points $s_n = (s_{n,1}, \cdots, s_{n,D})$, we calculate the *mean intra-class distances* for each class $C_l$

$$\bar{d}(C_l) = \frac{2}{N_l(N_l-1)} \sum_{i=1}^{N_l-1} \sum_{j=i+1}^{N_l} d\left(\mathbf{s}_i^{(l)}, \mathbf{s}_j^{(l)}\right),$$

and the *mean inter-class distances* for each pair of classes $C_l$ and $C_m$

$$\bar{d}(C_l, C_m) = \frac{1}{N_l N_m} \sum_{i=1}^{N_l} \sum_{j=1}^{N_m} d\left(\mathbf{s}_i^{(l)}, \mathbf{s}_j^{(m)}\right).$$

Here, $N_k$ is the number of points in class k, and $\mathbf{s}_i^{(k)}$ is the $i^{th}$ point of class k. The quantity $d(a, b)$ is the Euclidean distance between a and b. Finally, the Generalized Discrimination Value (GDV) is calculated from the mean intra-class and inter-class distances as follows:

$$\text{GDV} = \frac{1}{\sqrt{D}}\left[\frac{1}{L}\sum_{l=1}^{L}\bar{d}(C_l) - \frac{2}{L(L-1)}\sum_{l=1}^{L-1}\sum_{m=l+1}^{L}\bar{d}(C_l, C_m)\right]$$

whereas the factor $\frac{1}{\sqrt{D}}$ is introduced for dimensionality invariance of the GDV with D as the number of dimensions.

Note that the GDV is invariant with respect to a global scaling or shifting of the data (due to the z-scoring), and also invariant with respect to a permutation of the components in the N-dimensional data vectors (because the Euclidean distance measure has this symmetry). The GDV is zero for completely overlapping, non-separated clusters, and it becomes more negative as the separation increases. A GDV of -1 signifies already a very strong separation.

### *2.5. Code implementation:*

We collected the internal representations by passing the samples of text through the models. The process was implemented with Python, using PyTorch (Paszke et al., 2019) from HuggingFace's Transformers library (Wolf et al., 2020). Mathematical operations, like GDV calculations, were performed them with NumPy (Harris et al., 2020) and sci-kit-learn (Pedregosa et al., 2011) libraries. Visualizations and MDS were realised with Matplotlib (Hunter, 2007).

### 3. Results:

The neural model has different representations for linguistic constructions across its layers. We projected the model outputs for all linguistic constructions through the verbs' word embedding vectors using MDS (Figure 1 and 2). The GDV across layers of the neural network is illustrated for different verb constructions for each category: "agree" verbs, "come" verbs, and "give" verbs, and also for the value for all inputs. Consistent with the approach for analysing construction category representations, the GDV is used here to quantify the data separability and the strength of the clustering of verb representations within the models.

Principally, in the original BERT model as the layers progress from early to middle, there is a general trend of increased strong clustering for the within-category clustering, as indicated by the GDV moving away from 0.00 to -1.00. The "agree" verbs category, starting at a GDV of -0.062 and -0.105, showing week clustering at the 1st and 2nd layers respectively. Therefore, progresses to a most clustered value of -0.256 at the 4th layer. The "come" verbs, show a similar trend. However, clustered stronger than the "agree" verbs set exhibits initial weak clustering with a GDV value of -0.090, becoming more defined at -0.236 and -0.191 by layers 3rd and 4th respectively. The "give" verbs begin with a GDV value of -0.098 and achieve the strongest clustering with a GDV of -0.305 and -0.288 at the 3rd and 4th layers respectively.

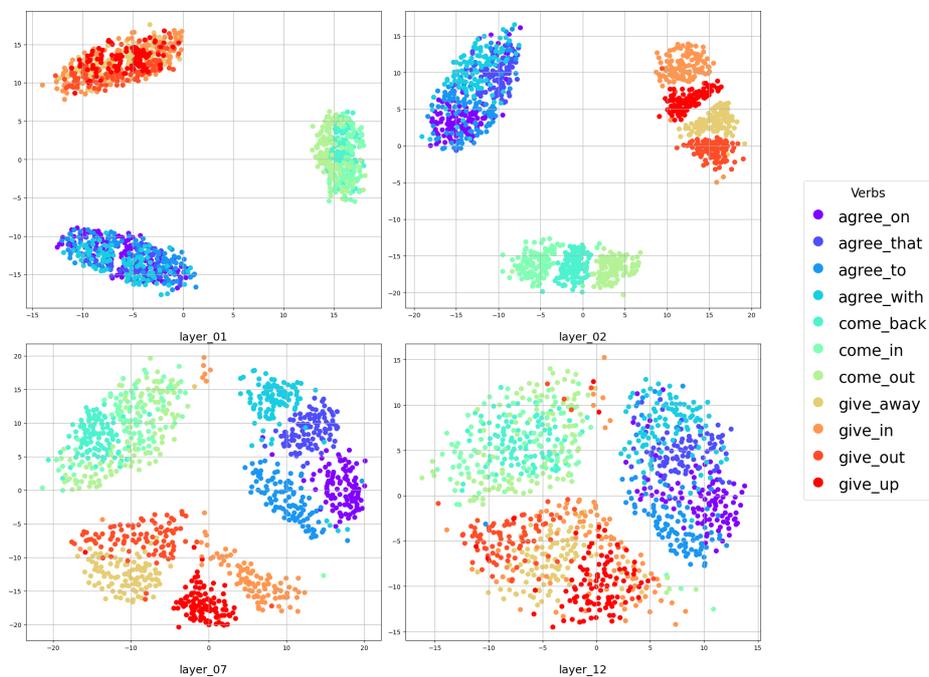

**Figure 1:** Early layers results of neural network testing and projection onto a 2-dimensional plane using MDS, with colour coding according to subsequent

construction. The points are strongly clusters based on the verb forms without considering the following particle in the 2$^{nd}$ layer for the phrasal verb categories *"come"* and *"give"*. However, no strong clustering for the category *"agree"*. The same is true for the 7$^{th}$ layer in the case of the verb *give* where still there is a clustering within the category based on the following particle. Moreover, there is clustering within the categories in the verb groups *"give"* and *"come"*.

The CxG-BERT showed the same trends, with variabilities of strong and weak clustering in different layers than the Original BERT. The "agree" category started with very weak clustering in the 1$^{st}$ layer with GDV of -0.017, while the string clustering recorded at the 7$^{th}$ layer with GDV of -0.162. The "come" verbs reported GDV of -0.039 in the 1$^{st}$ layer and the strongest clustering at the 7$^{th}$ layer with GDV of -0.100. The "give" verbs also started with weak clustering with GDV of -0.041. However, the 6$^{th}$ and 7$^{th}$ layers reported mostly the similar strong clustering with GDV of -0.153 and -0.151 Respectively.

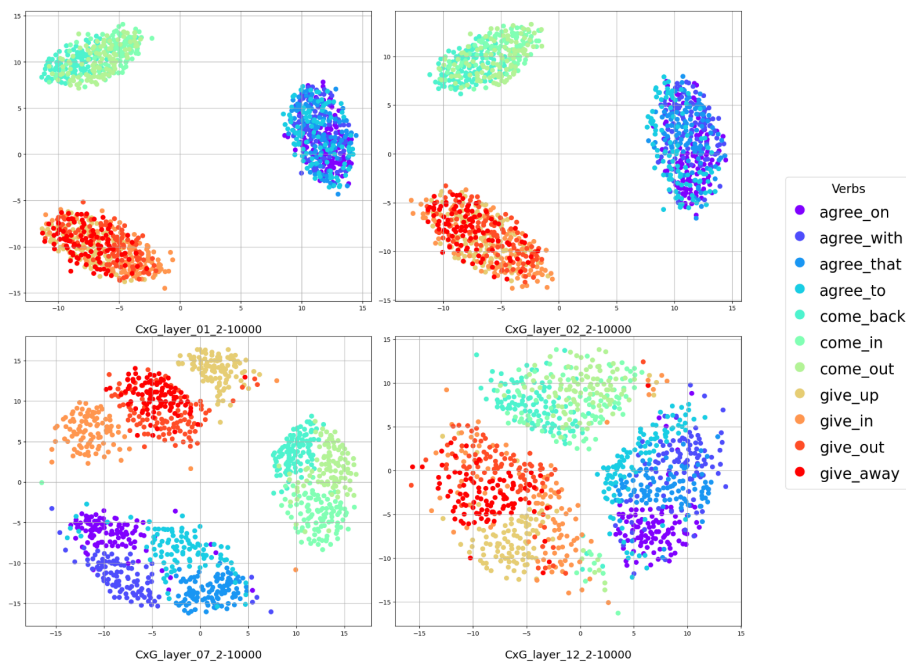

**Figure 2:** Layers output of neural network testing and projection onto a 2-dimensional plane using MDS, with colour coding according to subsequent constructions. The points are strongly clustered based on the verb forms, and then a clustering within the category is shown in the intermediate layers (7$^{th}$). The verbs in the late layers are not strongly clustered as much as in the intermediate layers. Moreover, an increased number of outliers has appeared, specifically in the '*give*' category.

These patterns across the layers indicate that the network's ability to cluster verbs based on their representations improves at the middle layers as the information flows through successive layers. Moreover, the verbs were treated similarly in the very early and late layers in the internal representations of both models, which indicates that they share the same internal representations in these layers, in contrast to the middle layers, specifically the 3$^{rd}$ and the 4$^{th}$ layers in the original BERT and 6$^{th}$ and 7$^{th}$ in the CxG-BERT, which achieved the strongest clustering. Regarding the GDV values for the whole input data, it starts with -0.358 at the 1st layer and -0.401 at the 3rd layer in the Original BERT. Whereas in the CxG-BERT it is reported a GDV of -0.329 at 1$^{st}$ layer which is the strongest clustering.

The analysis of embeddings extracted from BERT for the verb 'give' mostly 'give out' in the late layers (**Figure 1**) revealed outlier clusters, even though they are categorised within the same category. This embedding distinction could be observed, particularly in the types of constructions associated with the verb in each cluster. These findings may highlight the semantic and syntactic variations in the usage of the verb across different contexts.

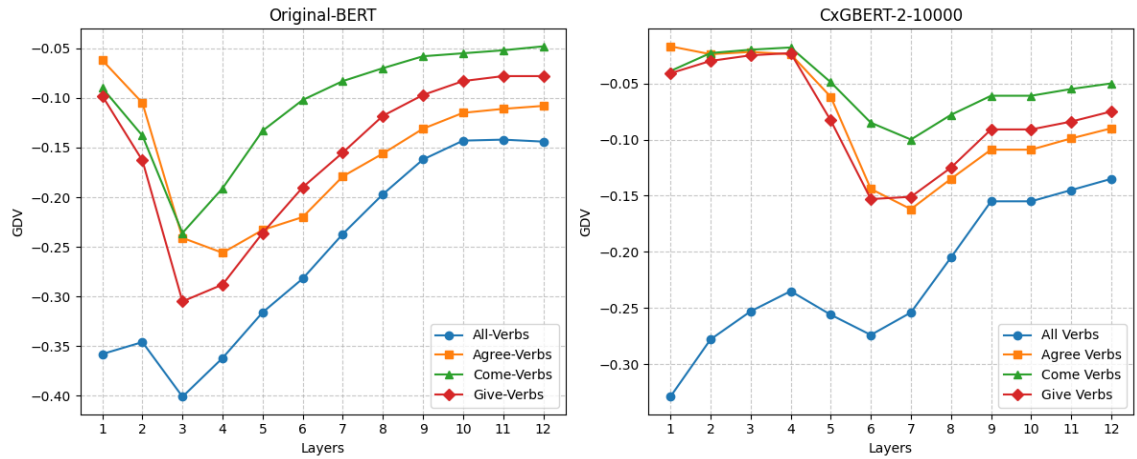

**Figure 3:** GDV curve across layers of the neural models. The decline of the GDV indicates that the neural network has representations of the different constructions that might oscillate between strong clustering between all the input data categories and strong clustering based on the verb forms. However, weak clustering was reported for the verbs within the category, as in the early and late layers, but not for the middle layers.

4. Discussion:

*Linguistic explanations:*

Considering the construction grammar framework, and the theoretical analysis of the phrasal verbs and verb-preposition combinations. We discuss the given results in terms of two approaches, the first is the lexical unity of phrasal verbs and the non-unity in verb-preposition combinations. The second is the representation of the constructions according to their organization in the constructicon network. As for the first approach, following Herbst and Schüller (2008: 120), we assumed the tokens involved in phrasal verb sequences to be represented in isolated spaces since they are already representing single lexical units in the lexicon (different representations of the token Give according to the phrasal verb it belongs to, and the same for the token

"come") since each phrasal verb has dependent meaning. However, the "agree" tokens should not be isolated in the dimensional space because they do not represent dependent meanings.

For the second analysis approach, (see tables 3 to 10 where the verbs are classified according to the full meaning of the constructions following the construct-i-con description by Herbst & Hoffman (2024). Thus, the different verbs in the category "give" are assumed to be represented differently according to the roles and meanings of the constructions they belong to (see the different meanings represented in tables 8/9/10), where the idiomatic nature of the phrasal verbs plays a role in their isolation of the representation of the meanings. The verbs under the category "come" share one constructional representation and also the token "come" has one meaning and the full meaning of the full verb particle construction is predicted from the combinations of its components.

For the category "agree", the same trend has been captured. However, in some levels of representation, it is not; such as the lower layer $2^{nd}$ where there was less clustering of the verbs within-category in comparison with other categories in the Original BERT. The weak clustering is also clear in several lower layers in the CxG-BERT. However, the "agree" verbs are similarly treated in comparison with other categories in the CxG-BERT.

Back to the results we see that the phrasal verbs belonging to the category "give" support the linguistic assumption on the lexical unity of phrasal verbs where we see that each verb is represented in isolated dimensional space in which the model is trained on a large amount of corpus data to acquire the syntactic and semantic properties of linguistic input (figure 1 and 2) this lexical unity is support interacted with the constructicon analysis of the semantic roles of the verbs as provided by

Herbst & Hoffman (2024). The same case of the phrasal verbs in the category "come" where we see them also represented in isolated dimensional space.

Taking into account the calculated GDV values, we explored the closest values to 0.00 which indicates weak clustering is achieved by the category "come" (Figure 2) in both models. We could align these GDV values with the constructional descriptions in (Table 7) where the three different verbs *come_in*, *come_back*, and *come_out* are all represented in the same constructicon. This might be explained by that the models capture the lexicosemantic information of the constructions from the single tokens. Therefore, even different verbs are presented they share same representations based on the contexts they appear in and also because their non-idiomaticity in which their meaning are predictable and the lexical verb participate with its meaning in the full meaning of verb-particle construction.

Earlier in this paper, we introduced the category "agree" as it supposed to show different behaviour than other categories since the theoretical framework in construction grammar has analysed the verb-preposition combination in different way than phrasal verbs Herbst and Schüller (2008). Therefore, we expected it to show less clustering since the following preposition does not play an important role in the understanding of the verb as the case of phrasal verbs—alternatively, in the case of *agree_on* and *agree_with*. However, we could not find any evidence to support this assumption.

***Models Capabilities:***

The analysis of linguistic representations within transformer-based models, such as BERT, highlights substantial differences across linguistic levels. Our study demonstrates that verb-particle combinations exhibit varying degrees of representation accuracy within different neural network layers. This variability underscores the complex interaction between neural architectures and linguistic structure, suggesting that the transformer's ability to model language depends significantly on the layer-specific features of its architecture.

One of the most striking findings from our study is the strong clustering observed in the early and intermediate layers of BERT, as opposed to the late layers. This pattern suggests that early to middle layers are crucial for capturing the core syntactic and semantic features of language, aligning with findings from previous studies that have emphasized the role of these layers in capturing the fundamental aspects of linguistic structure (Goldberg, 2019; Hewitt and Manning, 2019). These layers appear to form a 'sweet spot' where the neural representations are rich enough to distinguish between different linguistic constructions but not yet too specialized to the point of being overly task-specific.

In contrast, the weak clustering in the late layers indicates a lack of representation of semantic unity. This observation challenges the effectiveness of late layers in modeling cohesive semantic structures and suggests that these layers might be more influenced by the specific tasks the model was trained on, rather than general linguistic understanding. This finding diverges from traditional neurolinguistic methods, such as magnetoencephalography (MEG), which often reveal a strong coherence in semantic processing across the brain (Pulvermüller, 2002). The discrepancy raises questions about the extent to which deep learning models mimic human language processing and highlights the need for further research into how these models can be better aligned with neurobiological evidence.

The interpretation of outliers in our dataset further enriches our understanding of the model's limitations and capabilities. Outliers often represented cases where the standard semantic or syntactic rules did not apply, indicating areas where BERT struggles to form consistent representations. This aspect of the analysis is critical for improving model robustness and adapting neural networks to handle linguistic anomalies effectively.

Our findings resonate with and contribute to a broader body of research in computational linguistics and cognitive neuroscience. Studies such as those by Rogers et al. (2020) and Belinkov et al. (2020) have similarly explored the linguistic capabilities of neural models, often focusing on different aspects of language processing. By connecting our results with these past studies, we not only validate our methodologies but also extend the current understanding of how deep learning models can be utilized and improved for complex linguistic tasks.

In conclusion, our research underscores the capabilities of neural language models in processing linguistic sequences and provides a foundation for future investigations aimed at advancing the linguistic representational power of these systems. It also calls for a multidisciplinary approach, combining insights from computational linguistics, cognitive science, and neuroscience, to develop models that more accurately reflect the complexity of human language understanding.